\documentclass[11pt,a4paper]{article}
\usepackage[hyperref]{emnlp-ijcnlp-2019}
\usepackage{times}
\usepackage{latexsym}

\usepackage{url}
\usepackage{amssymb}
\usepackage{amsmath}
\usepackage{graphicx}
\usepackage{bm}
\usepackage{multirow}
\usepackage{makecell, diagbox}
\usepackage{subfigure}

\aclfinalcopy 

\title{Fine-grained Knowledge Fusion for Sequence Labeling Domain Adaptation}

\author{Huiyun Yang$^{1,2}$ \quad Shujian Huang$^{1,2}$ \quad Xinyu Dai$^{1,2}$ \quad Jiajun Chen$^{1,2}$ \\
National Key Laboratory for Novel Software Technology, Nanjing, China$^{1}$ \\
Nanjing University, Nanjing, China$^{2}$ \\
\texttt{yanghy@smail.nju.edu.cn} \\
\texttt{\{huangsj, daixinyu, chenjj\}@nju.edu.cn}\ \\
}

\date{}

\begin{document}
\maketitle
\begin{abstract}
In sequence labeling, previous domain adaptation methods focus on the adaptation from the source domain to the entire target domain without considering the diversity of individual target domain samples, which may lead to negative transfer results for certain samples. Besides, an important characteristic of sequence labeling tasks is that different elements within a given sample may also have diverse domain relevance, which requires further consideration. To take the multi-level domain relevance discrepancy into account, in this paper, we propose a fine-grained knowledge fusion model with the domain relevance modeling scheme to control the balance between learning from the target domain data and learning from the source domain model. Experiments on three sequence labeling tasks show that our fine-grained knowledge fusion model outperforms strong baselines and other state-of-the-art sequence labeling domain adaptation methods.

\end{abstract}

\section{Introduction}

Sequence labeling tasks, such as Chinese word segmentation (CWS), POS tagging (POS) and named entity recognition (NER), are fundamental tasks in natural language processing. Recently, with the development of deep learning, neural sequence labeling approaches have achieved pretty high accuracy \citep{chen2017,zhang2018NER}, relying on large-scale annotated corpora. However, most of the standard annotated corpora belong to the news domain, and models trained on these corpora will get sharp declines in performance when applied to other domains like social media, forum, literature or patents \citep{daume2007,Blitzer2007}, which limits their application in the real world. Domain adaptation aims to exploit the abundant information of well-studied source domains to improve the performance in target domains \citep{pan2010}, which is suitable to handle this issue. 
Following \citet{daume2007}, we focus on the supervised domain adaptation setting, which utilizes large-scale annotated data from the source domain and small-scale annotated data from the target domain. 

\begin{table}
\small
\centering
\begin{tabular}{|c|c|}
\hline
Types & Cases \\
\hline
\multirow{2}*{Strongly} & \emph{Ops} \textbf{Steve Jobs resigned as CEO of Apple.} \\
  & \textbf{Share prices are rising} \emph{soooo} \textbf{fast}! \\
\hline
\multirow{2}*{Weakly} & \emph{Alas} \textbf{as time goes by}, \emph{hair's gone}. \\
  & \emph{Rock to 204 Section} \textbf{next week}! \\
\hline
\end{tabular}
\caption{Tweets from the social media domain have different degrees of relevance to the source domain (news). Within each case, the bold part is strongly relevant and the italic part is weakly relevant.}

\label{table0}
\end{table}

For sequence labeling tasks, each sample is usually a sentence, which consists of a sequence of words/Chinese characters, denoted as the element.
We notice an interesting phenomenon: different target domain samples may have varying degrees of domain relevance to the source domain. As depicted in Table \ref{table0}, there are some tweets similar to the news domain (i.e. strongly relevant). But there are also some tweets of their own style, which only appear in the social media domain (i.e. weakly relevant). The phenomenon can be more complicated for the cases where the whole sample is strongly relevant while contains some target domain specific elements, or vice versa, showing the diversity of relevance at the element-level. In the rest of this paper, we use `domain relevance' to refer to the domain relevance to the source domain, unless specified otherwise.


\begin{figure}[t]
\centering
\subfigure[Previous methods]{
\label{(a)}
\includegraphics[width=0.23\textwidth]{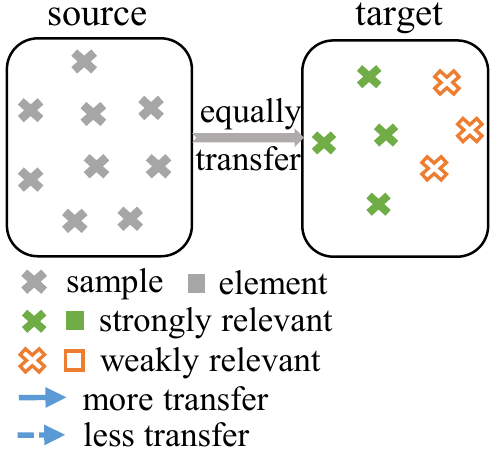}}
\subfigure[Our method]{
\label{(b)} 
\includegraphics[width=0.23\textwidth]{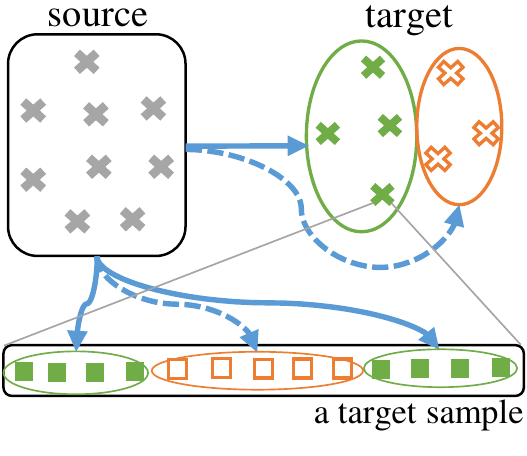}}
\caption{
Previous methods transfer knowledge by the whole sample set, while our method consider diverse domain relevance within the target domain set and within every target sample to transfer knowledge respectively.}
\label{fig0}
\end{figure}

Conventional neural sequence labeling domain adaptation methods \cite{liu2012,liu2014,zhang2014,chen2017,peng2017,lin2018} mainly focus on reducing the discrepancy between the sets of source domain samples and target domain samples. However, they neglect the diverse domain relevance of individual target domain samples, let alone the element-level domain relevance. As depicted in Figure \ref{fig0}, obviously, strongly relevant samples/elements should learn more knowledge from the source domain, while weakly relevant samples/elements should learn less and keep their characteristics. 

In this paper, we propose a fine-grained knowledge fusion model to control the balance between learning from the target domain data and learning from the source model, inspired by the knowledge distillation method \cite{Bucila2006,Hinton2015}.
With both the sample-level and element-level domain relevance modeling and incorporating, the fine-grained knowledge fusion model can alleviate the \emph{negative transfer} \cite{Rosenstein2005} in sequence labeling domain adaptation.


 We verify the effectiveness of our method on six domain adaptation experiments of three different tasks, i.e. CWS, POS and NER, in two different languages, i.e. Chinese and English, respectively. Experiments show that our method achieves better results than previous state-of-the-art methods on all tasks. We also provide detailed analyses to study the knowledge fusion process.


Contributions of our work are summarized as follows:
\begin{itemize}
\item We propose a fine-grained knowledge fusion model to balance the learning from the target data and learning from the source model.
\item We also propose multi-level relevance modeling schemes to model both the sample-level and element-level domain relevance. 
\item Empirical evidences and analyses are provided on three different tasks in two different languages, which verify the effectiveness of our method.
\end{itemize}

\section{Knowledge Distillation for Adaptation}
\label{basic ts}
Knowledge distillation (KD), which distills the knowledge from a sophisticated model to a simple model, has been employed in domain adaptation \cite{bao2017,meng2018}.
Recently, online knowledge distillation\cite{Furlanello2018,zhou2018} is shown to be more effective, which shares lower layers between the two models and trains them simultaneously.

\begin{figure}[t]
\centering
\includegraphics[scale=0.6]{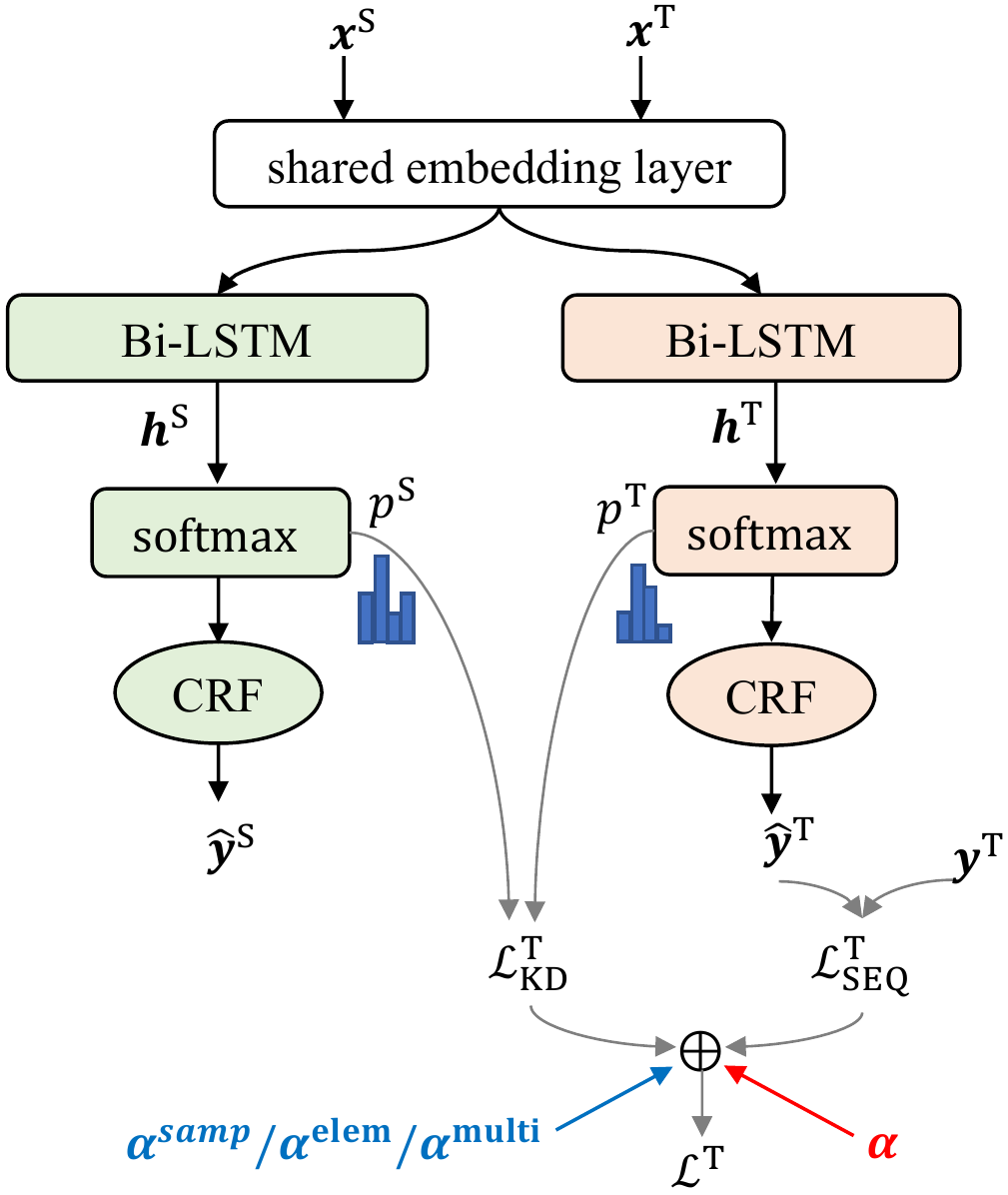}
\caption{The architecture of basicKD (with the red $\alpha$, see \S \ref{basic ts}) or fine-grained knowledge fusion model (with the blue $\alpha$, see \S \ref{fine-grained}), where the green part belongs to the source model, the orange part belongs to the target model and the white part is common. Better viewed in color.
}
\label{fig1}
\end{figure}

For sequence labeling domain adaptation, we utilize the online knowledge distillation method to distill knowledge from the source model to improve the target model, denoted as basicKD, which is depicted in Figure \ref{fig1}.
We use the Bi-LSTM-CRF architecture~\cite{Huang2015}, for both the source model and the target model, and share the embedding layer between them. 
  
 
\noindent\textbf{Notations} \quad For the rest of the paper, we use the superscript $S$ and $T$ to denote the source domain and the target domain, respectively. 
Source domain data is a set of $m$ samples with gold label sequences, denoted as $(\mathbf{x}_{j}^{S}, \mathbf{y}_{j}^{S})_{j=1}^m$. Similarly, target domain data has $n$ samples, denoted as $(\mathbf{x}_{i}^{T}, \mathbf{y}_{i}^{T})_{i=1}^n$, where $n \ll m$.

The training loss of the source model is the cross entropy between the predicted label distribution $\hat{\mathbf{y}}$ and the gold label $\mathbf{y}$:
\begin{equation}
    \mathcal{L}^{S} = -\frac{1}{m} \sum_{j=1}^m \mathbf{y}^S_{j}\log \hat{\mathbf{y}}^S_{j}
\end{equation}

The training loss of the target model is composed of two parts, namely the sequence labeling loss $\mathcal{L}^{T}_{\text{ SEQ}}$ and the knowledge distillation loss $\mathcal{L}^{T}_{\text{KD}}$:

\begin{equation}
    \mathcal{L}^{T} = (1-\alpha)\mathcal{L}^T_{\text{SEQ}} + \alpha \mathcal{L}^T_{\text{KD}}
\end{equation}
\begin{equation}
    \mathcal{L}^{T}_{\text{SEQ}} = -\frac{1}{n} \sum_{i=1}^n \mathbf{y}^T_{i}\log\hat{\mathbf{y}}^T_{i}
\end{equation}
\begin{equation}
    \mathcal{L}^{T}_{\text{KD}} = -\frac{1}{n} \sum_{i=1}^n \mathbf{p}^S_{i}\log{\mathbf{p}}^T_{i}
\end{equation}
where $\mathcal{L}^T_{\text{SEQ}}$ is similar to $\mathcal{L}^{S}$, while $\mathcal{L}^T_{\text{KD}}$ is the cross entropy between the probability distributions  predicted by the source model and the target model. $\alpha$ is a hyper-parameter scalar, which is used to balance the learning from the target domain data and the learning from the source model.

\section{Relevance Modeling}
\label{relevance}
BasicKD provides individual learning goals for every sample and element of the target domain, using a scalar $\alpha$ to weight. As a result, the source model has the same influence on all target samples, in which the diversity of domain relevance is neglected.  

\begin{figure}[t]
\centering
\includegraphics[scale=0.6]{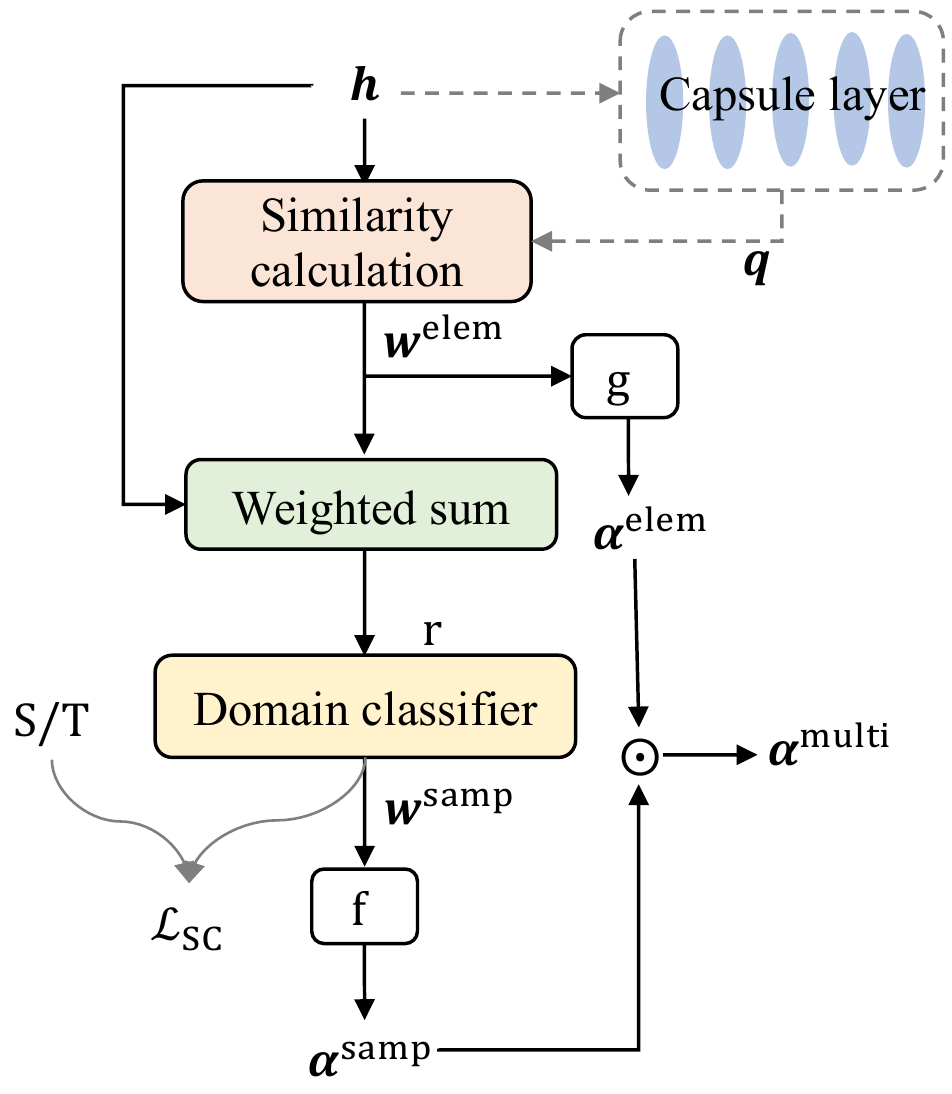}
\caption{The relevance modeling process (see \S \ref{relevance}), where the block f denotes Eq.(\ref{eqsamp}) and the block g denotes Eq.(\ref{eqelem}). 
}
\label{fig2}
\end{figure}

Here we present methods to model the domain relevance of target samples and elements, which could then be used to guide the knowledge fusion process (see \S \ref{fine-grained}). The overall architecture is shown in Figure \ref{fig2}. The relevance of each sample is a scalar, denoted as the sample-level relevance weight, $\mathbf{w}^\text{samp}_i$ for the $i^{th}$ sample, which can be obtained by the sample-level domain classification. The relevance of each element is also a scalar, while the relevance weights of all elements within a sample form a weight vector $\mathbf{w}^\text{elem}$, which can be obtained by the similarity calculation.



\subsection{Element-level Relevance}
To acquire the element-level relevance, we employ the domain representation $\bm{q} \in \mathbb{R}^{2d_h}$ ($d_h$ is the dimension of the Bi-LSTM) and calculate the similarity between the element representation and the domain representation. We incorporate two methods to get $\bm{q}$: (1) Domain-q: $\bm{q}$ is a trainable domain specific vector, where every element within a domain share the same $\bm{q}$; (2) Sample-q: $\bm{q}$ is the domain relevant feature extracted from each sample, where every element within a sample share the same $\bm{q}$. 
Because of the superiority of the capsule network modeling abstract features \cite{gong2018,yang2018},
we use it to capture the domain relevant features within a sample. We incorporate the same bottom-up aggregation process as \citet{gong2018} and the encoded vector is regarded as $\bm{q}$:
\begin{equation}
    \mathbf{q} = \text{Capsule}(\mathbf{h})
\end{equation}
where $\mathbf{h}$ is the hidden state matrix of a sample.

The similarity calculation formula is the matrix dot \footnote{We also try dot and MLP, while matrix dot get better performance with fewer parameters.}:
\begin{equation}
    \mathbf{w}_{j}^\text{elem} = \mathbf{q}^\top \mathbf{B} \mathbf{h}_{j}
\end{equation}
where $\bm{h}_{j}$ is the hidden states of the $j^{\text{th}}$ element and $\mathbf{w}_{j}^\text{elem}$ is the relevance weight of it. $\mathbf{B} \in \mathbb{R}^{2d_h \times 2d_h}$ is a trainable matrix.

\subsection{Sample-level Relevance}
To acquire the sample-level domain relevance, we make use of the domain label to carry out sample-level text classification (two class, source domain or target domain).
The weight $\mathbf{w}^\text{elem}$ is normalized across the sample length using the softmax function, then the sample representation can be obtained by the weighted sum of hidden states. The process can be expressed as:
\begin{equation}
    \hat{\mathbf{w}}_{j}^\text{elem} = \frac{\exp(\mathbf{w}_{j}^\text{elem})}{\sum_{k}\exp(\mathbf{w}_{k}^\text{elem})}
\end{equation}
\begin{equation}
    \mathbf{r} = \sum_{j=1}^{L}\hat{\mathbf{w}_{j}}^\text{elem} \cdot \mathbf{h}_{j}
\end{equation}
$\mathbf{r} \in \mathbb{R}^{2d_h}$ is the sample representation and $L$ is the sample length.

Once the sample representation is obtained, the multi-layer perceptron (MLP) and softmax do sample classification next:
\begin{equation}
    [w^{\text{samp}}, 1- w^{\text{samp}} ] = {[\text{softmax}(\text{MLP}(\mathbf{r}))]}^\top
\end{equation}
where $w^{\text{samp}}$ is the sample relevance weight.

\section{Fine-grained Knowledge Fusion for Adaptation}
\label{fine-grained}
With the relevance modeling, the fine-grained knowledge fusion model is proposed to fuse the knowledge from the source domain and the target domain at different levels. The overall architecture is shown in Figure \ref{fig1}.

\subsection{Sample-level Knowledge Fusion}
Different samples of target domain tend to show different domain relevance, and as a result, they need to acquire different amount of knowledge from the source domain. 
Different $\alpha$ is assigned to each target sample based on its domain relevance to achieve the sample-level knowledge fusion. The new $\alpha$ can be computed as:
\begin{equation}
    \bm{\alpha}^{\text{samp}}_i = \sigma(\tau \cdot \mathbf{w}_i^{\text{samp}}+\gamma)
\label{eqsamp}
\end{equation}
where $\bm{\alpha}^{\text{samp}}_i$ is the $\alpha$ of the $i^{\text{th}}$ sample and $\mathbf{w}_i^{\text{samp}}$ is the relevance weight of it; $\sigma$ denotes the sigmoid function; $\tau$ is temperature and $\gamma$ is bias.

The loss functions of the target model can be computed as:
\begin{equation}
\label{eq9}
    \mathcal{L}^{T} = \mathcal{L}^T_{\text{SEQ}} + \mathcal{L}^T_{\text{KD}}
\end{equation}
\begin{equation}
\label{eq7}
    \mathcal{L}^T_{\text{SEQ}} = -\frac{1}{n} \sum_{i=1}^n (1-\bm{\alpha}^{\text{samp}}_i) \mathbf{y}^T_{i}\log\hat{\mathbf{y}}^T_{i}
\end{equation}
\begin{equation}
\label{eq8}
    \mathcal{L}^T_{\text{KD}} = -\frac{1}{n} \sum_{i=1}^n \bm{\alpha}^{\text{samp}}_i \mathbf{p}^S_{i} \log{\mathbf{p}}^T_{i}
\end{equation}

The sample classification losses of the source model $\mathcal{L}_{sc}^S$ and target model $\mathcal{L}_{sc}^T$ are both cross entropy.

\subsection{Element-level Knowledge Fusion}
Besides the sample-level domain relevance, different elements within a sample tend to present diverse domain relevance. 
In this method, we assign different $\alpha$ to each element based on its domain relevance weight to achieve the element-level knowledge fusion. The new $\alpha$ can be computed as:
\begin{equation}
    \bm{\alpha}^{\text{elem}}_i = \sigma(\mathbf{W}_{\alpha}\mathbf{w}^{\text{elem}}_i+\mathbf{b}_{\alpha})]
\label{eqelem}
\end{equation}
where $\bm{\alpha}^{\text{elem}}_i \in \mathbb{R}^{L}$ is a vector, in which $\bm{\alpha}^{\text{elem}}_{ij}$ denotes the $\alpha$ of the $j^{\text{th}}$ element in the $i^{\text{th}}$ sample. $\mathbf{w}^{\text{elem}}_i$ is the relevance weight of the $i^{\text{th}}$ sample. $\mathbf{W}_{\alpha}$ and $\mathbf{b}_{\alpha}$ are trainable parameters. 

\begin{table*}[t]
\footnotesize
\centering
\begin{tabular}{|c|c|c|c|c|}
\hline
Task & Language & Source & Target & Domain \\
\hline
\multirow{2}*{CWS} & Chinese & CTB6 \cite{xue2005} & Zhuxian \cite{zhang2014} & news $\to$ novels \\
\cline{2-5}
 & Chinese & CTB6 \cite{xue2005} & Weibo \cite{qiu2016overview} & \multirow{5}*{news $\to$ social media} \\
\cline{1-4}
\multirow{2}*{POS} & Chinese & CTB6 \cite{xue2005} & Weibo \cite{qiu2016overview} & \\
\cline{2-4}
 & English & PTB \cite{PTB} &  Twitter \cite{twitterpos} &  \\
\cline{1-4}
\multirow{2}*{NER} & Chinese & MSRA \cite{levow2006} & WeiboNER \cite{peng2015} & \\
\cline{2-4} 
 & English & Ontonotes \cite{ontonotes} & Twitter \cite{twitterpos} & \\
\hline
\end{tabular}
\caption{Datasets used in this paper.}
\label{table1}
\end{table*}


The loss functions of the target model can be expressed as:
\begin{equation}
\label{eq13}
    \mathcal{L}^T_{\text{SEQ}} = -\frac{1}{n} \sum_{i=1}^n \sum_{j=1}^L (1-\bm{\alpha}^{\text{elem}}_{ij}) \mathbf{y}^T_{ij}\log\hat{\mathbf{y}}^T_{ij}
\end{equation}
\begin{equation}
\label{eq14}
    \mathcal{L}^T_{\text{KD}} = -\frac{1}{n} \sum_{i=1}^n \sum_{j=1}^L \bm{\alpha}^{\text{elem}}_{ij} \mathbf{p}^S_{ij} \log{\mathbf{p}}^T_{ij}
\end{equation}
where $*_{ij}$ denotes the $*$ of the $j^{\text{th}}$ element in the $i^{\text{th}}$ sample, and the final loss function is the same with Eq.(\ref{eq9}).

\begin{table}[t]
\footnotesize
\centering
\begin{tabular}{l}
\hline
$\mathbf{Algorithm\ 1}$ Training Process of Knowledge Fusion \\
\hline
1. $\mathbf{Input}$: source data, target data\\
2. $\mathbf{Hyper-parameters}$: batch size $b$, teach step $I$ \\
3. Initialize parameters of the source and target model \\
4. (optional) Use the source data to pre-train $\bm{\theta}^S$ and $\bm{\theta}^T$  \\
5. $\mathbf{repeat}$ \\
6. \ \qquad $\mathbf{for}$ $i=1$ to $I$ $\mathbf{do}$ \\
7. \ \qquad \qquad Sample $b$ samples from the source data \\
8. \ \qquad \qquad Compute $\mathcal{L}^S$, and update $\bm{\theta}^S$ \\
9. \ \qquad \qquad Compute $\mathcal{L}_{sc}^S$, and update $\bm{\theta}^S$ \\
10. \qquad $\mathbf{end\ for}$ \\
11. \qquad Use $\bm{\theta}^S$ to test $\mathbf{x}_{train}^T$ and get $\mathbf{p}^S$ \\
12. \qquad $\mathbf{while}$ in an episode: \\
13. \qquad \qquad Sample $b$ samples from the target data \\
14. \qquad \qquad Use relevance modeling to get $\mathbf{w}^\text{samp}$,$\mathbf{w}^\text{elem}$ \\
15. \qquad \qquad Compute $\bm{\alpha}^{\text{samp}}$/$\bm{\alpha}^{\text{elem}}$/$\bm{\alpha}^{\text{multi}}$ and $\mathcal{L}^T_{\text{SEQ}}$ \\
16. \qquad \qquad Use $\bm{\theta}^T$ to predict $\mathbf{p}^T$, and compute $\mathcal{L}^T_{\text{KD}}$ \\
17. \qquad \qquad Compute $\mathcal{L}^T$, and update $\bm{\theta}^T$ \\
18. \qquad \qquad Compute $\mathcal{L}_{sc}^T$, and update $\bm{\theta}^T$ \\
19. \qquad $\mathbf{end\ while}$ \\
20. $\mathbf{until}$ converge \\
\hline
\end{tabular}
\label{alg1}
\end{table}

\subsection{Multi-level Knowledge Fusion}
In this method, we take both the sample-level and element-level relevance diversities into account to implement the multi-level knowledge fusion, and the multi-level $\alpha$ can be computed as:
\begin{equation}
    \bm{\alpha}^{\text{multi}} = \bm{\alpha}^{\text{samp}}\odot\bm{\alpha}^{\text{elem}}
\end{equation}
where $\odot$ denotes the element-wise product. $\bm{\alpha}^{\text{multi}} \in \mathbb{R}^{n \times L}$ is a matrix as well.

The loss functions of the target model can be obtained by replacing $\bm{\alpha}^{\text{elem}}_{ij}$ with $\bm{\alpha}^{\text{multi}}_{ij}$ in Eq.(\ref{eq13}) and Eq.(\ref{eq14}).

\subsection{Training Process}
Both the source model and the target model can be pre-trained on the source domain data (warm up, optional). In the fine-grained knowledge fusion method, the source model and the target model are trained alternately. Within an episode, we use $I$ steps to train the source model ahead, then the soft target ($\mathbf{p}^S$) can be obtained and the target model will be trained. During the training of the target model, the parameters of the source model are fixed (gradient block). Every training step includes the sequence labeling training and the sample classification training. We conduct early stopping according to the performance of the target model. The whole training process is shown in Algorithm 1.

\section{Experiments}

\subsection{Datasets}
We conduct three sequence labeling tasks: CWS, POS and NER, and the latter two tasks containing both Chinese and English settings. Detailed datasets are shown in Table \ref{table1}. There are two kinds of source-target domain pairs: news-novels and news-social media. To be consistent with the setting where there is only small-scale target domain data, we use 5\% training data of Weibo for both CWS and POS. For the different NER tag sets,  we only focus on three types of entities: Person (PER), Location (LOC) and Organization (ORG) and regard other types as Other (O).

\subsection{Settings}
For each task, hyper-parameters are set via grid search on the target domain development set.
Embedding size and the dimension of LSTM hidden states is set to 100. Batch size is set to 64. Learning rate is set to 0.01. We employ the dropout strategy on the embedding and MLP layer with the rate of 0.2. The $l_2$ regularization term is set to 0.1. The gradient clip is set to 5. The teach step $I$ is set to 100. The routing iteration is set to 3 and the number of the output capsules is set to 60. The temperature $\tau$ is initialized to 1 and the probability bias $\gamma$ is initialized to 0.5. We set the $\alpha$ of the basicKD method to 0.5 according to \citet{Hinton2015}. 
We randomly initialize the embedding matrix without using extra data to pre-train, unless specified otherwise. 

\begin{table*}[!htbp]
\small
\centering
\begin{tabular}{|c|c|c|c|c|c|c|c|c|}
\hline
\multirow{3}{*}{Methods} & \multicolumn{4}{c|}{CWS} & \multicolumn{2}{c}{\multirow{2}*{POS}} & \multicolumn{2}{|c|}{\multirow{2}*{NER}} \\
\cline{2-5}
 & \multicolumn{2}{c|}{Zhuxian} & \multicolumn{2}{c|}{5\% Weibo} & \multicolumn{2}{c}{} & \multicolumn{2}{|c|}{}\\
\cline{2-9}
  & F & $R_{\text{OOV}}$ & F & $R_{\text{OOV}}$ & zh & en & zh & en \\
\hline
Target only & 92.80 & 65.81 & 84.01 & 64.12 & 93.03 & 86.83 & 46.49 & 59.58 \\
BasicKD & 94.23 & 74.08 & 89.21 & 76.26 & 95.69 & 89.96 & 49.92 & 62.15 \\
\hline
Pre-trained embedding & 93.70 & 70.44 & 87.62 & 72.27 & 94.96 & 89.70 & 52.53 & 61.36 \\
Pre-trained model & 94.43 & 74.30 & 89.50 & 76.27 & 96.10 & 90.05 & 54.25 & 62.88 \\
Linear projection & 94.14 & 72.75 & 88.77 & 75.85 & 95.92 & 89.36 & 52.71 & 62.27 \\
Domain mask & 94.30 & 75.20 & 88.84 & 75.03 & 96.01 & 89.81 & 54.12 & 62.64 \\
NAL & 94.47 & 74.62 & 88.63 & 75.77 & 96.19 & 90.48 & 54.70 & 63.32 \\
AMCL  & 94.62 & 74.46 & 89.42 & 76.16 & 94.13 & 89.12 & 51.47 & 61.57 \\
\hline
FGKF & 95.01 & 77.26 & 90.45 & 77.27 & $\bm{96.60}$ & 91.33 & 55.60 & 63.81\\
+ Pre-trained embedding & $\bm{95.09}$ & $\bm{77.56}$ & $\bm{90.73}$ & $\bm{77.87}$ & 96.36 & $\bm{91.66}$ & $\bm{57.57}$ & $\bm{65.51}$ \\
\hline
\end{tabular}
\caption{Results of domain adaptation on three tasks, where zh denotes the Weibo datasets (in Chinese), and en denotes the Twitter dataset (in English).}
\label{table3}
\end{table*}

\subsection{Baselines}

We implement several baseline methods, including: \textbf{source only} (training with only source domain data), \textbf{target only} (training with only target domain data) and \textbf{basicKD} (see \S \ref{basic ts}). 

We also re-implement state-of-the-art sequence labeling domain adaptation methods, following their settings except for unifying the embedding size and the dimension of LSTM hidden states: 

\begin{itemize}
    \item \textbf{Pre-trained methods:} \textbf{Pre-trained embedding} incorporates source domain data with its gold label to pre-train context-aware character embedding \cite{zhou2017}, which is used to initialize the target model; \textbf{Pre-trained model} trains the model on the source domain and then finetune it on the target domain.
    \item \textbf{Projection methods:} \textbf{Linear projection} \cite{peng2017} uses the domain-relevant matrix to transform the learned representation from different domains into the shared space; \textbf{Domain mask} \cite{peng2017} masks the hidden states of Bi-LSTM to split the representations into private and public regions to do the projection; \textbf{Neural adaptation layer} (NAL) \cite{lin2018} incorporates adaptation layers at the input and output to conduct private-public-private projections.
    \item \textbf{Adversarial method:} \textbf{Adversarial multi-criteria learning} (AMCL) \cite{chen2017} uses the shared-private architecture with the adversarial strategy to learn the shared representations across domains.
\end{itemize}

\subsection{Overall Results on CWS}
We use the F1-score (F) and the recall of out-of-vocabulary words (R$_{\text{oov}}$) to evaluate the domain adaptation performance on CWS. We compare methods with different relevance modeling schemes and different levels of knowledge fusion, without warm up. And we denote our final model as \textbf{FGKF}, which is the multi-level knowledge fusion with the sample-q relevance modeling and warm up. 

\begin{table}[ht]
\small
\centering
\begin{tabular}{|c|c|c|c|c|}
\hline
\multirow{2}{*}{Methods} & \multicolumn{2}{c|}{Zhuxian} & \multicolumn{2}{c|}{5\% Weibo} \\
\cline{2-5}
 & F & $R_{\text{OOV}}$ & F & $R_{\text{OOV}}$ \\
\hline
Source only & 83.86 & 62.40 & 83.75 & 70.74 \\
Target only & 92.80 & 65.81 & 84.01 & 64.12 \\
BasicKD & 94.23 & 74.08 & 89.21 & 76.26 \\
\hline
Domain-q $\bm{\alpha}^{\text{samp}}$ & 94.55 & 74.02 & 89.63 & 75.93 \\
Domain-q $\bm{\alpha}^{\text{elem}}$ & 94.81 & 74.75 & 89.99 & $\bm{77.59}$ \\
Domain-q $\bm{\alpha}^{\text{multi}}$ & 94.75 & 74.96 & 90.06 & 77.25 \\
\hline
Sample-q $\bm{\alpha}^{\text{samp}}$ & 94.57 & 74.47 & 89.77& 76.81 \\
Sample-q $\bm{\alpha}^{\text{elem}}$ & 94.78 & 74.52 & 90.07 & 76.94 \\
Sample-q $\bm{\alpha}^{\text{multi}}$ & 94.91 & 75.56 & 90.20 & 77.46 \\
\hline
FGKF & $\bm{95.01}$ & $\bm{77.26}$ & $\bm{90.45}$ & 77.27 \\
\hline
\end{tabular}
\caption{Results of baselines and fine-grained knowledge fusion methods on CWS.}
\label{table2}
\end{table}

The results in Table \ref{table2} show that both the basicKD method and fine-grained methods achieve performance improvements through domain adaptation. Compared with the basicKD method, FGKF behaves better (+1.1\% F and +2.8\% R$_{\text{oov}}$ v.s.  basicKD on average), as it takes multi-level relevance discrepancies into account. The sample-q method performs better than the domain-q method, which shows the domain feature is better represented at the sample level, not at the domain level. As for the granularity of $\alpha$, the performances of $\bm{\alpha}^{\text{elem}}$ is better than $\bm{\alpha}^{\text{samp}}$, showing the necessity of modeling element-level relevance.  
And there isn't a distinct margin between $\bm{\alpha}^{\text{elem}}$ and $\bm{\alpha}^{\text{multi}}$ as most of the multi-level domain relevance can be included by the element level. 
Results of FGKF with warm up indicate that starting from sub-optimal point is better than starting from scratch for the target model.

Among related works (Table \ref{table3}), AMCL and Pre-trained model methods have better performances in CWS. Compared with other methods, FGKF achieves the best results in both F and R$_{\text{OOV}}$. 
Results demonstrate the effectiveness of our fine-grained knowledge fusion architecture for domain adaptation, and also show the significance of considering sample-level and element-level relevance discrepancies.

\begin{figure*}[ht]
\centering
\vspace{-0.2cm}
\includegraphics[scale=0.58]{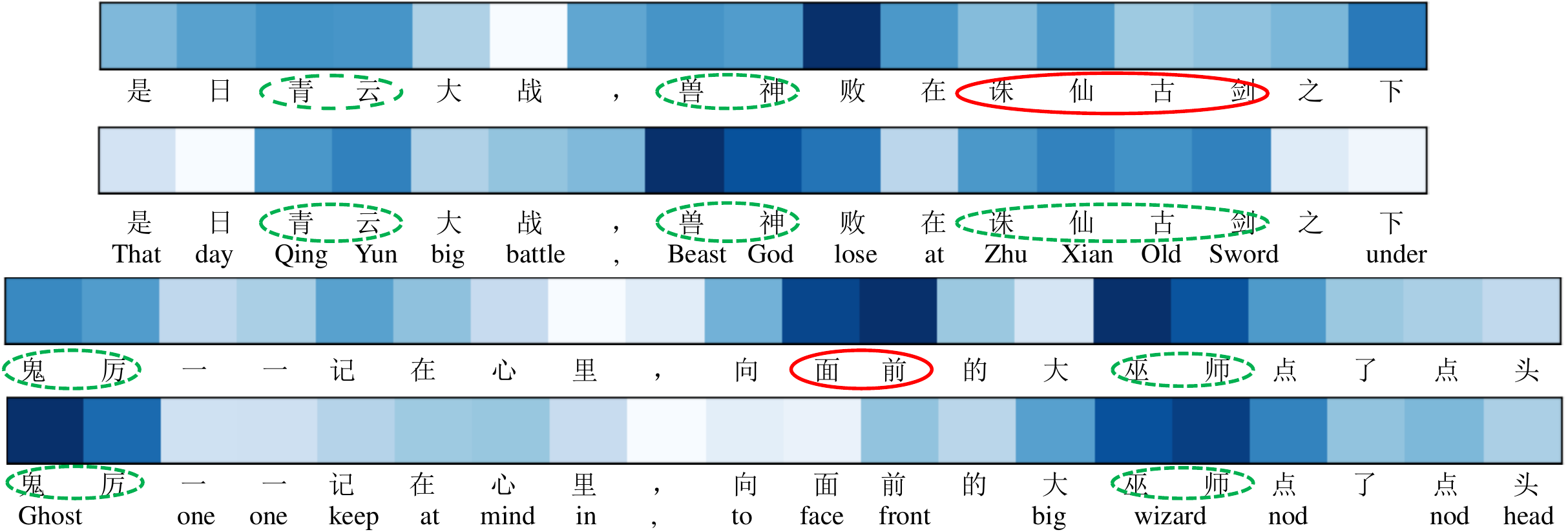}
\vspace{-0.1cm}
\caption{Two cases of the element-level relevance modeling visualization, where the upper one belongs to the domain-q method and the lower one belongs to the sample-q method. The green dotted circle indicates the correct domain relevant element and the red solid circle indicates the ignored or mistaken extracted element.}
\label{fig3}
\end{figure*}

\subsection{Overall Results on POS and NER}
To further verify the effectiveness of FGKF, we conduct experiments on POS and NER tasks, using F1-score as the evaluation criterion. Detailed results are shown in Table \ref{table3}. In these tasks, FGKF achieves better results than other adaptation methods. Extra gain could be obtained by using pre-trained embedding. These results also verify the generalization of our method over different tasks and languages.

\section{Analysis}
In this section, we will display and discuss the domain adaptation improvements provided by our fine-grained knowledge fusion method.

\subsection{Performances of Elements with Different Relevance}
To further probe into the experimental results of the fine-grained knowledge fusion, we classify the target test data (in element level) into two classes: strongly relevant and weakly relevant, based on their relevance degrees to the source domain. The partition threshold is according to the average relevance score of the target training data. Detailed results on Twitter are depicted in Table \ref{tabel7}.

\begin{table}[!htbp]
\footnotesize
\centering
\begin{tabular}{|c|c|c|c|c|}
\hline
\multirow{2}{*}{Methods} & \multicolumn{2}{c}{POS} & \multicolumn{2}{|c|}{NER} \\
\cline{2-5}
 & Strong & Weak & Strong & Weak \\
\hline
Source only & 87.47 & 82.48 & 68.27 & 46.30 \\
Target only & 86.46 & 87.41 & 62.01 & 56.29 \\
BasicKD & 91.92 & \underline{83.82} & 70.20 & \underline{52.63} \\
FGKF & $\bm{92.55}$ & $\bm{89.93}$ & $\bm{71.81}$ & $\bm{57.92}$ \\
\hline
\end{tabular}
\caption{Results of the strongly/weakly relevant elements on the Twitter test set.}
\label{tabel7}
\end{table}

It is reasonable that both the basicKD and FGKF enhance the performance of the strongly relevant part, while FGKF get larger improvements because it is able to enhance the knowledge fusion by learning more from the source model. For the weakly relevant part, the basicKD method damages the performance on it (from 87.41 to 83.82 for POS and from 56.29 to 52.63 for NER), which indicate the negative transfer. On the contrary, FGKF improves the performance of the weakly relevant part compared with the target only baseline with a large margin. It is shown that the fine-grained domain adaptation method can reduce the negative transfer on the weakly relevant part and contribute to the transfer on the strongly relevant one.

\subsection{Relevance Weight Visualization}
We carry out the visualization of the element-level relevance weight to illustrate the effects of the two relevance modeling schemes (domain-q and sample-q). Figure \ref{fig3} exhibits two cases of element-level relevance modeling results, from which we can explicitly observe that the two schemes capture different domain relevance within a sample.
In the first case, the sample-q method extracts more domain relevant elments, like ``Qingyun", ``Beast God" and ``Zhuxian Old Sword", while the domain-q method ignores the last one. In the second case, the domain-q method extracts ``front" incorrectly. These results indicate that the sample-q method can implement better relevance modeling than the domain-q method to some extent, and prove that the domain relevant feature is better represented at the sample level, not at the domain level.

\begin{table*}[!htbp]
\footnotesize
\centering
\begin{tabular}{|c|c c c c c|c c c c c|}
\hline
Tasks & \multicolumn{5}{c|}{POS} & \multicolumn{5}{c|}{NER}\\
\hline
Sentence & I & got & u & next & week & Louis & interview & with & The & Sun \\
\hline
Source only & PN & VBD & \underline{NN} & JJ & NN & B-PER & O & O & \underline{O} & \underline{O} \\
Target only & PN & \underline{VBZ} & PN & JJ & NN & \underline{O} & O & O & B-ORG & I-ORG \\
BasicKD & PN & VBD & \underline{NN} & JJ & NN & B-PER & O & O & \underline{O} & \underline{O} \\
FGKF & PN & VBD & PN & JJ & NN & B-PER & O & O & B-ORG & I-ORG \\
\hline 
\end{tabular}
\caption{Two cases of domain adaptation, where the underlined tags are wrong.}
\label{tabel9}
\end{table*}

\subsection{Case Study}
We take two samples in Twitter test set as examples to show how the element-level relevance affects the adaptation. Results in Table \ref{tabel9} show that both basicKD and FGKF can improve the performance of strongly relevant elements, e.g. ``got (VBD)", ``Lovis (B-PER)". However, only FGKF reduces the transfer of source domain errors, e.g. ``u (NN)", ``The (B-ORG) Sun (I-ORG)".

\subsection{Ablation Study}
We conduct the ablation study on Twitter dataset (Table \ref{tabel8}). Results show the  gradient block and the multi-level knowledge fusion are of vital importance to FGKF. The embedding sharing and warm up also make contributions.

\begin{table}[!htbp]
\footnotesize
\centering
\begin{tabular}{|c|c|c|c|c|}
\hline
\multirow{2}{*}{Methods} & \multicolumn{2}{c|}{POS} & \multicolumn{2}{c|}{NER} \\
\cline{2-5}
 & F & $\Delta$ & F & $\Delta$ \\
\hline
FGKF & 91.33 & - & 63.81 & - \\
\hline
w/o share embedding & 90.75 & -0.58 & 62.47 & -1.34 \\
w/o gradient block & 88.48 & -2.85 & 58.83 & -4.98 \\
w/o $\bm{\alpha}^{\text{samp}}$ & 90.94 & -0.39 & 63.52 & -0.30 \\
w/o $\bm{\alpha}^{\text{elem}}$ & 90.23 & -1.10 & 62.43 & -1.38 \\
w/o $\bm{\alpha}^{\text{multi}}$ & 90.12 & -1.21 & 62.32 & -1.49 \\
w/o warm up & 90.89 & -0.44 & 63.17 & -0.64 \\
\hline
\end{tabular}
\caption{Ablation results of the Twitter test set.}
\label{tabel8}
\end{table}

\subsection{Influence of Target Data Size}
\begin{figure}[ht]
\centering
\includegraphics[scale=0.45]{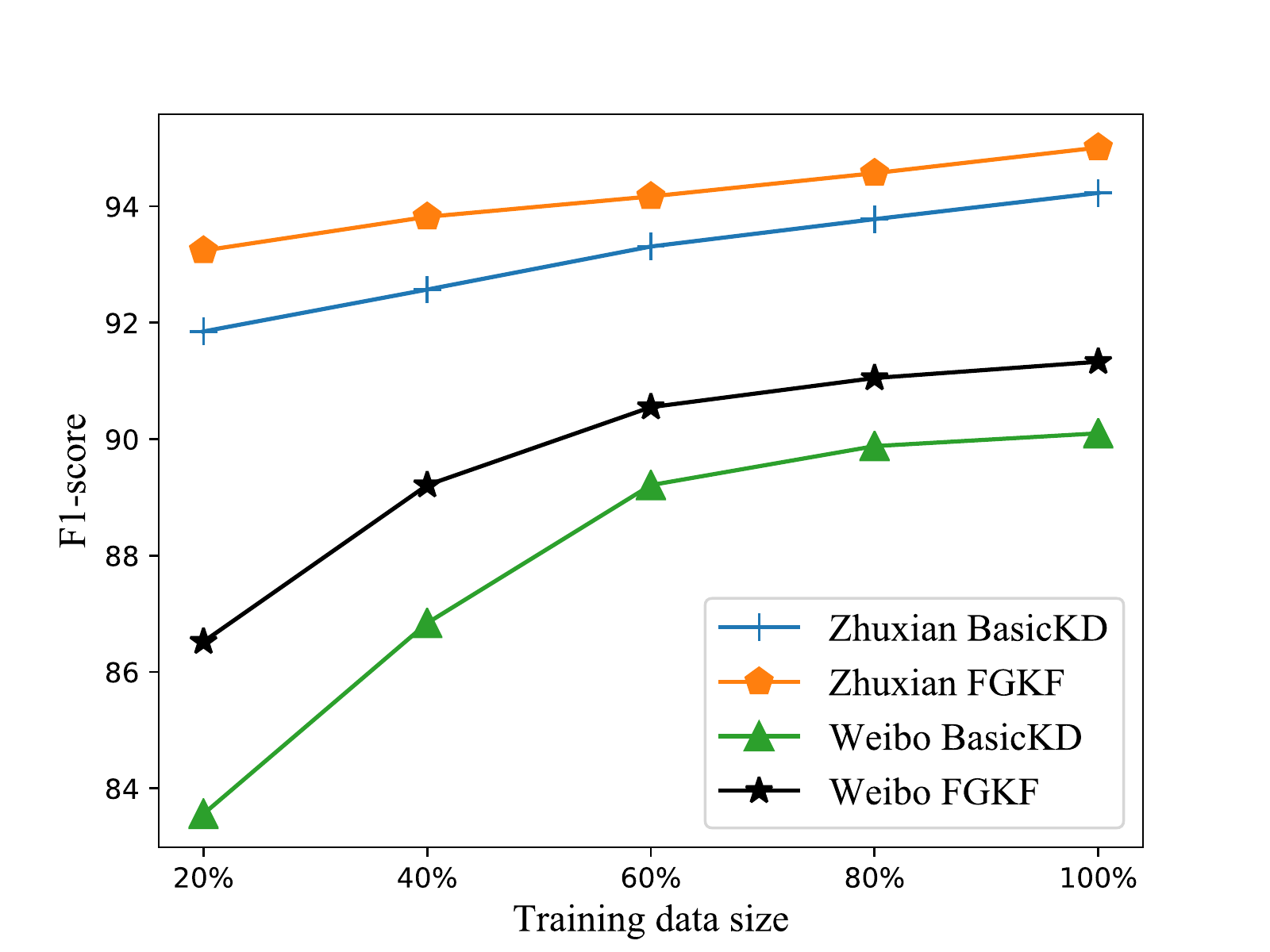}
\caption{Results of CWS target test set with varying target training data size. Only 10\% training data of Weibo is utilized.}
\label{fig4}
\end{figure}

Here we investigate the impact of the target domain data size on FGKF. As is depicted in Figure \ref{fig4}, when the size is small (20\%), the gap is pretty huge between FGKF and basicKD, which verifies the significance of fine-grained knowledge fusion in the low-resource setting. Even with the size of target data increasing, there are still stable margins between the two methods.

\section{Related Work}



Besides the source domain data, some methods utilize the target domain lexicons \cite{liu2014,zhang2014}, unlabeled \cite{liu2012} or partial-labeled target domain data \cite{liu2014} to boost the sequence labeling adaptation performance, which belong to unsupervised or semi-supervised domain adaptation. However, we focus on supervised sequence labeling domain adaptation, where huge improvement can be achieved by utilizing only small-scale annotated data from the target domain.


Previous works in domain adaptation often try to find a subset of source domain data to align with the target domain data \cite{chopra2013,Ruder2017} which realizes a kind of source data sample or construct a common feature space, while those methods may wash out informative characteristics of target domain samples. 
Instance-based domain adaptation \cite{jiang2007,zhang2018} implement the source sample weighting by assigning higher weights to source domain samples which are more similar to the target domain.
There are also some methods \cite{guo18moe,kim2017,zeng2018} explicitly weighting multiple source domain models for target samples in multi-source domain adaptation.
However, our work focuses on the supervised single source domain adaptation, which devote to implementing the knowledge fusion between the source domain and the target domain, not within multiple source domains.
Moreover, considering the important characteristics of sequence labeling tasks, we put more attention to the finer-grained adaptation, considering the domain relevance in sample level and element level.

\section{Conclusion}
In this paper, we propose a fine-grained knowledge fusion model for sequence labeling domain adaptation to take the domain relevance diversity of target data into account. With the relevance modeling on both the sample level and element level, the knowledge of the source model and target data can achieve multi-level fusion. Experimental results on different tasks demonstrate the effectiveness of our approach, and show the potential of our approach in a broader range of domain adaptation applications. 

\section*{Acknowledgements}
We would like to thank the anonymous reviewers for their insightful comments. Shujian Huang is the corresponding author. This work is supported by National Science Foundation of China (No. U1836221, No. 61772261), National Key R\&D Program of China (No. 2019QY1806).

\bibliography{emnlp}
\bibliographystyle{acl_natbib}

\end{document}